\documentclass[runningheads]{llncs}

\usepackage[T1]{fontenc}
\usepackage[dvipsnames]{xcolor}
\usepackage[inline]{enumitem}
\usepackage[font=small,labelfont=bf,tableposition=top]{caption}
\usepackage{amsmath}
\usepackage{graphicx}
\usepackage{bm}
\usepackage{multirow}
\usepackage{cuted}
\usepackage{comment}
\usepackage{subcaption}  
\usepackage{booktabs}

\newcommand\tstrut{\rule{0pt}{2.6ex}}         
\newcommand\bstrut{\rule[-0.9ex]{0pt}{0pt}}   
\newcommand\improv{\small\textcolor{ForestGreen}}  
\usepackage{wrapfig}

\definecolor{cvprblue}{rgb}{0.21,0.49,0.74}
\usepackage[pagebackref,breaklinks,colorlinks,citecolor=cvprblue]{hyperref}

\begin{document}

\title{Harlequin: Color-driven Generation of Synthetic Data for Referring Expression Comprehension}
\titlerunning{Harlequin: Color-driven Generation of Synthetic Data for REC}

\author{
  Luca Parolari\inst{1}\orcidID{0000-0001-8574-4997} \and
  Elena Izzo\inst{1}\orcidID{0000-0003-1105-4162} \and
  Lamberto Ballan\inst{1}\orcidID{0000-0003-0819-851X}
}

\institute{
  University of Padova, Italy \\
  \email{\{luca.parolari,elena.izzo\}@phd.unipd.it} \\
  \email{lamberto.ballan@unipd.it}
}

\maketitle

\begin{abstract}
Referring Expression Comprehension (REC) aims to identify a particular object in a scene by a natural language expression, and is an important topic in visual language understanding.
State-of-the-art methods for this task are based on deep learning, which generally requires expensive and manually labeled annotations. Some works tackle the problem with limited-supervision learning or relying on Large Vision and Language Models. However, the development of techniques to synthesize labeled data is overlooked. 
In this paper, we propose a novel framework that generates artificial data for the REC task, taking into account both textual and visual modalities. 
At first, our pipeline processes existing data to create variations in the annotations. 
Then, it generates an image using altered annotations as guidance. The result of this pipeline is a new dataset, called \emph{Harlequin}, made by more than 1M queries. 
This approach eliminates manual data collection and annotation, enabling scalability and facilitating arbitrary complexity.
We pre-train three REC models on Harlequin, then fine-tuned and evaluated on human-annotated datasets. Our experiments show that the pre-training on artificial data is beneficial for performance.

\end{abstract}

\section{Introduction}

The expressiveness and variety of the human language are the basis of communication between people. Their ability to interact and understand each other attracts researchers to design models able to communicate with them. In this context, the task of Referring Expression Comprehension (REC)~\cite{DBLP:conf/cvpr/Yu0SYLBB18}, also known as Visual Grounding~\cite{li2021referring,DBLP:conf/iccv/DengYCZL21} or Phrase Localization~\cite{grounder16,DBLP:conf/iccv/WangS19a}, aims to identify a specific object in a scene described by a phrase, called referring expression or sometimes query. The research progress in this task has been made possible thanks to the active development of datasets. Since 2015, Flickr30k Entities~\cite{DBLP:conf/iccv/PlummerWCCHL15}, ReferIt~\cite{DBLP:conf/emnlp/KazemzadehOMB14}, RefCOCO, and two variants RefCOCO+ and RefCOCOg,~\cite{DBLP:conf/eccv/YuPYBB16,DBLP:conf/cvpr/MaoHTCY016} were released.
\begin{wrapfigure}{r}{0.5\textwidth}
    \centering
    \includegraphics[width=1\linewidth]{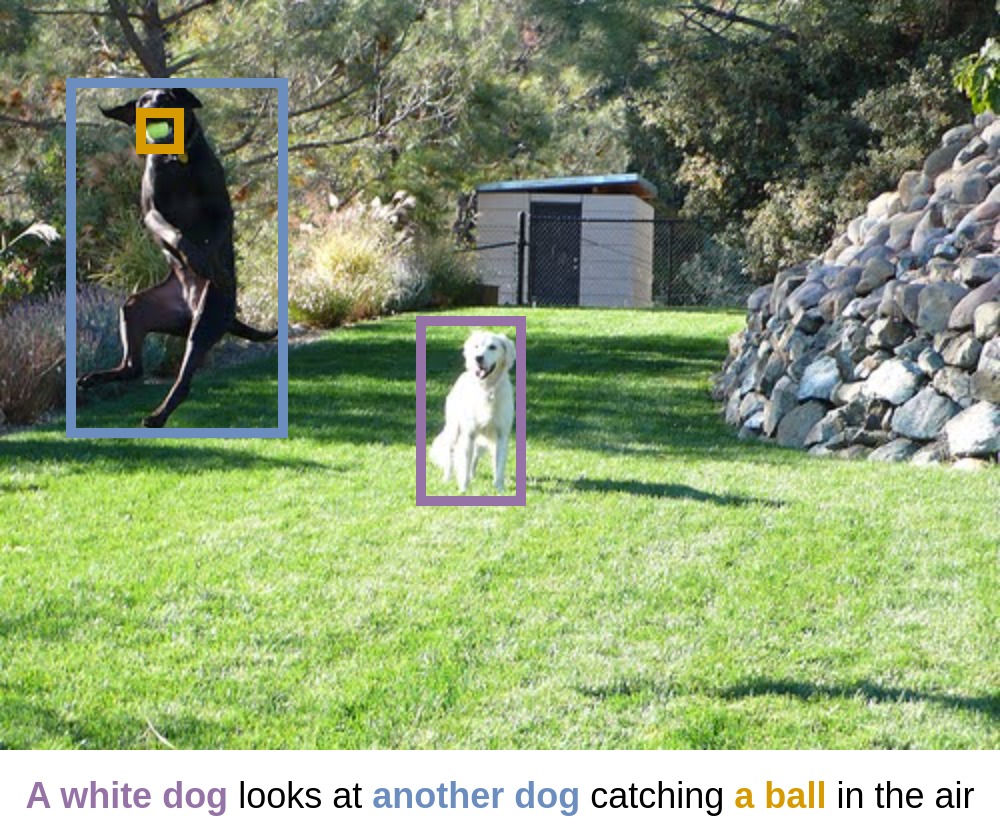}
    \caption{Annotations required by the Referring Expression Comprehension task. In this example, the image has one caption with three referring expressions. Each referring expression is accompanied by the location of the referred object (bounding box).}
    \label{fig:rec-task}
\end{wrapfigure}
These datasets are human-labeled and consist of triplets composed of an image, a referring expression, and a bounding box. Fig.~\ref{fig:rec-task} shows an example. However, the gathering and annotation of such data is time-consuming and resource-intensive, representing a critical bottleneck for the collection of sufficiently large training sets and new benchmarks.

Current works face this issue exploring limited supervision learning techniques such as weakly-supervised \cite{DBLP:conf/bmvc/RigoniPSSB23}, semi-supervised~\cite{DBLP:conf/icassp/JinYLH23}, and unsupervised~\cite{DBLP:conf/iccv/WangS19a} or rely on large Vision and Language models pre-trained on a massive amount of multimodal data~\cite{DBLP:conf/iccv/KamathSLSMC21}. However, the development of techniques and pipelines to create new, reasoning-oriented datasets is overlooked, limited by fine-grained annotations required by the Referring Expression Comprehension task. Some works explore the generation of the queries by either working on their properties or structure ~\cite{DBLP:conf/iccv/TanakaINSUH19,DBLP:conf/cvpr/ChenWMWW20,DBLP:conf/cvpr/JiangLHSH22}.
However a method to generate both queries and images has not been investigated yet.

In this paper, we propose a pipeline for generating synthetic data for the Referring Expression Comprehension task, taking into account both textual and visual modalities. Recent developments in text-to-image generation with diffusion models allowed fine-grained control over the output by either embedding guidance signals like bounding box, keypoints, or semantic maps with language~\cite{DBLP:conf/cvpr/LiLWMYGLL23} or even expressing them by means of text~\cite{DBLP:conf/eccv/YangGW000LW22}.
Inspired by these advancements, we argue that 
\begin{enumerate*}[label=(\roman*)] \item the process of manual collection and annotation of data for this task can finally be avoided, and \item new benchmarks with arbitrary size and complexity can be created. The proposed pipeline and extensive experiments we run address those hypotheses.\end{enumerate*}

Broadly speaking, our pipeline is composed of two modules. The first is the Annotation Generation Engine. It is responsible for generating new referring expressions (REs) with consistent bounding box annotations. We use REs from Flickr30k Entities as seeds and generate their variations to keep consistency with the arrangement of objects in the image. REs are altered by varying their attributes, specifically the color attribute. The second is the Image Generation Engine. Guided by the annotation obtained in the previous step, it generates a new image. The synthesized image should represent the given caption and depict objects at specific locations that look like the given description. Objects are described through referring expressions, which may have varied attributes.

Following this strategy, we synthetically generate Harlequin, a new dataset consisting of train, validation, and test sets. Harlequin is the first dataset totally synthetic generated for the Referring Expression Comprehension task.
The experiments show that its use in pre-training stage boost the results on real data, reducing labeling effort and errors in annotations.

Our contributions can be summarized as follows:
\begin{enumerate*}[label=(\roman*)]
    \item We propose a novel pipeline for generating synthetic data for the Referring Expression Comprehension task, increasing richness and variability and reducing to zero the human effort required for collecting annotations; 
    \item We introduce Harlequin, a new dataset for the Referring Expression Comprehension task, which is entirely synthetically generated;
    \item We prove the effectiveness of our synthetic dataset if used in a pre-training stage to transfer knowledge on real datasets;
    \item We release both the dataset and the code.\footnote{\href{https://github.com/lparolari/harlequin}{https://github.com/lparolari/harlequin}}
\end{enumerate*}

\section{Related Work}
\label{sec:related}

\paragraph{Referring Expression Comprehension} Among different approaches studied in literature~\cite{DBLP:conf/cvpr/Yu0SYLBB18,DBLP:conf/iccv/YangGWHYL19}, recently the transformer-based approach emerged, demonstrating superior performance. TransVG~\cite{DBLP:conf/iccv/DengYCZL21} makes use of transformer for both intra- and inter-modality correspondence. VLTVG~\cite{DBLP:conf/cvpr/YangXYLLH22} employs a language-guided context encoder to extract discriminative features of the referred object. QRNet~\cite{DBLP:conf/cvpr/YeTYYWZ0L22} introduces query-aware dynamic attention to extract query-refined visual features with a hierarchical structure. VG-LAW~\cite{DBLP:conf/cvpr/Su0DW0L023} adds adaptive weights to the visual backbone to make it an expression-specific feature extractor. LGR-NET~\cite{DBLP:journal/tcsv/LuLFMW24} emphasizes the guidance of the referring expression for cross-modal reasoning. InterREC~\cite{DBLP:journals/tmm/WangPXS23} increases object-level relational-level interpretability through an image semantic graph and a reasoning order tree.

\paragraph{Synthetic Data Generation} 
In the last decade different areas of research started to investigate the use and generation of synthetic data to lower the cost of data and automation collection. 
In~\cite{DBLP:journals/corr/GaidonWCV16}, the authors argued the interchangeability between real and synthetic datasets and demonstrated the improvements of performance pre-training the models on virtual data encouraging the generation of synthetic data in various domains such as autonomous driving~\cite{DBLP:journals/corr/GaidonWCV16},
gardening~\cite{DBLP:conf/wacv/LeMDKG21}, deepfake detection~\cite{DBLP:journals/corr/abs-2304-00500},
object detection and 3D reconstruction~\cite{DBLP:journals/corr/abs-2208-04052}. In many cases, datasets were created by means of simulators which guarantee complete control over synthetic environments such as Unity~\cite{DBLP:journals/corr/abs-1809-02627}, Blender~\cite{Blender} and CARLA~\cite{DBLP:conf/corl/DosovitskiyRCLK17}. Newer trends instead employ generative models to increase the automation in data generation and labeling process~\cite{DBLP:journals/corr/abs-2304-00500}.

\paragraph{Text-to-image Generation} Diffusion-based models demonstrated astonishing abilities in generating complex and realistic images. Recently, the existing pre-trained text-to-image diffusion models allowed fine-grained control in image generation, specifying requirements at the level of bounding boxes, masks, and edge or depth maps. For example, GLIGEN~\cite{DBLP:conf/cvpr/LiLWMYGLL23} uses a pre-trained T2I diffusion model and, freezing its weights, injects the grounding information into new trainable layers via a gated mechanism, focusing primarily on bounding boxes as the grounding condition. Similarly, ReCo~\cite{DBLP:conf/cvpr/YangWGLLWD0L0W23} extends stable diffusion, adding position tokens to enable open-ended regional texts for high-level region control. Finally, ControlNet~\cite{DBLP:conf/iccv/ZhangRA23} introduces conditional control connecting the trainable copy and the large pre-trained text-to-image diffusion models via ``zero convolution'' layers to eliminate harmful noise during training.

\section{Human-annotated Datasets for Referring Expression Comprehension}
\label{sec:motivation}

The most popular datasets in the field are Flickr30k Entities, ReferIt, and especially RefCOCO family. All these datasets were built on top of existing sets of captioned images, and they were human-annotated to align a referring expression with the bounding box of the mentioned entity in the image. In particular,  Flickr30k Entities dataset was built to augment Flickr30k~\cite{DBLP:journals/tacl/YoungLHH14} image captions with 244k coreference chains yielding almost 276k bounding boxes in 32k images. ReferIt, which contains images from the TC-12 expansion of the ImageCLEF IAPR dataset~\cite{DBLP:journals/cviu/EscalanteHGLMMSPG10}, has 131k expressions in 20k photographs of natural scenes. RefCOCO, RefCOCO+, and RefCOCOg were built on top of MSCOCO dataset~\cite{DBLP:conf/eccv/LinMBHPRDZ14}. RefCOCO and RefCOCO+ count 142k referring expressions in 20k images, instead RefCOCOg counts 85k expressions in 27k images. They aimed to collect images with multiple instances of the same object class to increase the complexity.
Besides, RefCOCOg focused on rich and natural descriptions, while RefCOCO and RefCOCO+ 
on appearance-based descriptions. 

Although crowd-sourcing protocols allowed the collection of a noticeable amount of annotations, we believe that such a time-consuming and resource-intensive task severely limits the gathering of new datasets where generalization, adaptability, and reasoning properties can be learned and evaluated.
In this paper, we investigate a pipeline for generating synthetic data for the Referring Expression Comprehension task, having control of both visual and textual content.
As a starting point towards this direction, we decide to add some constraints in the generation of the data in order to properly validate the pipeline, the synthetic dataset and its applicability to real data. In particular, inspired by~\cite{DBLP:journals/corr/GaidonWCV16}, we generate a dataset applying variations on the existing Flickr30k Entities one. Among others, we chose Flickr30k Entities for seed samples because every image is annotated with a sentence, yielding many referring expressions. From a generative point of view, this setting alleviates the amount of guessing and constrains the possible space of images that can be generated to a subset, where objects are precisely described and spatially located.

\section{The Proposed Pipeline}
\label{sec:method}

\begin{figure*}[ht]
    \centering
    \includegraphics[width=1\linewidth]{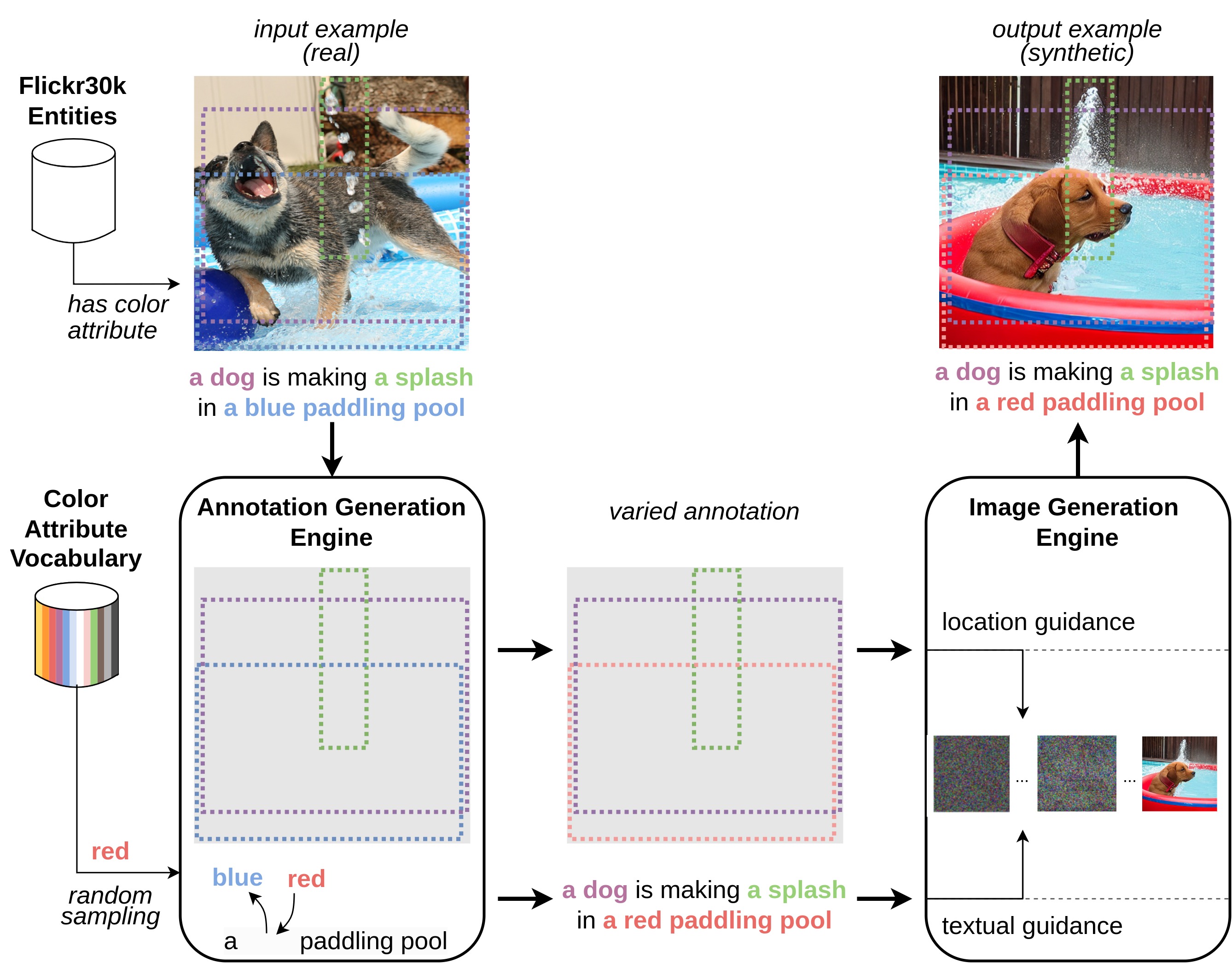}
    \caption{Our pipeline. It processes existing samples from Flickr30k Entities data. We select the ones characterized by at least one \textit{color} attribute in their referring expressions. The Annotation Generation Engine processes the sample's caption, referring expressions and locations where the color attribute is replaced with a randomly chosen color. The caption is updated accordingly. Then, the Image Generation Engine creates the new image using new annotations provided by the Annotation Generation Engine as guidance for the generation.}
    \label{fig:method}
\end{figure*}

The proposed approach, depicted in Fig.~\ref{fig:method}, relies on two components to generate synthetic data for Referring Expression Comprehension. The former, termed \textbf{Annotation Generation Engine}, is in charge of creating annotations to guide image generation. The latter, named \textbf{Image Generation Engine}, is responsible for synthesizing images enforcing the guidance provided by the Annotation Generation Engine. Specifically, given an input annotation $\bm{a}$ composed of an image caption $\bm{c}$ and set of referring expressions along with referred object locations $\{(\bm{q}_i, \bm{l}_i)\}_{i=1}^{N}$, the Annotation Generation Engine produces new annotations by varying attributes in the $p$-th referring expression, with $p \in [1, N]$. Then, the Image Generation Engine uses the annotation provided by the Annotation Generation Engine to generate a synthetic image \textit{I} exploiting GLIGEN~\cite{DBLP:conf/cvpr/LiLWMYGLL23}, a generative model based on Stable Diffusion~\cite{DBLP:conf/cvpr/RombachBLEO22}.

Since people frequently use colors to describe and disambiguate objects~\cite{DBLP:conf/emnlp/KazemzadehOMB14}, we select \textit{\textbf{color}} as the attribute to alter in the annotations. 
The color attribute has also proven to have a strong impact in several computer vision tasks, ranging from visual recognition problems (like object detection and image captioning) \cite{lu2024} to visual tracking \cite{danelljan2014}.
Therefore, for each referring expression we generate several variations where the color attribute is replaced with a new color. This enhance richness and variability of the dataset, because with one single seed many other examples can be generated representing objects with different colors, and possibly new orientations, perspectives and views of the same scene.
Moreover, altering the \textit{color} attribute offers some advantages:
\begin{enumerate*}[label=(\roman*)]
    \item a simple variation of the textual content has a strong impact on the generated images, allowing the models to learn to disambiguate between similar scenes;
    \item this alteration does not affect the position of the object in the image, retaining the original layout of objects in the image;
    \item it is (relatively) easy to understand and manipulate by a generative model.
\end{enumerate*}

\subsection{The Annotation Generation Engine}
\label{sec:age}

The Annotation Generation Engine (AGE) is a function defined over the set of annotations $\mathcal{A}$. It is specifically designed for Referring Expression Comprehension task and produces compatible annotations by altering queries in existing samples: $\phi: \mathcal{A} \rightarrow \mathcal{A}$. The AGE component takes an annotation $\bm{a}$ in input. The annotation consists of a caption $\bm{c}$ and a non-empty set of entities $E$. Each entity is described by the textual form of a referring expression and the location of the referred object:
\begin{align}
    \text{Annotation: } & \bm{a} = (\bm{c}, E) \\
    \text{Caption: } & \bm{c} = [c_1, \cdots, c_L] \\
    \text{Entities: } & E = \{(\bm{q}_i, \bm{l}_i)\}_{i=1}^N
\end{align}
where $\bm{c}$ is a caption of $L$ tokens, $N$ is the number of referring expressions, $\bm{q_i} = [c_j, \ldots, c_k]$ with $1 \leq j \leq k \leq L$ is the textual representation of the referring expression from a subset of contiguous tokens in $\bm{c}$, $\bm{l_i} = [\alpha_{\text{min}}, \beta_{\text{min}}, \alpha_{\text{max}}, \beta_{\text{max}}]$ is with top-left and bottom-right coordinates of the referred object. The AGE returns a new annotation where the $p$-th referring expression is varied by replacing a color attribute, $p \in [1, N]$. The location is not altered. Tokens in the caption are updated accordingly to the new referring expression, while other referred objects are not varied and serve as context. Mathematically, the output of $\phi(\bm{a})$ is $\bm{\hat{a}} = (\bm{\hat{c}}, \hat{E})$ where
\begin{align}
    \bm{\hat{c}} & = [c_1, \cdots, c_{j-1}, \overbrace{\hat{c}_j, \cdots, \hat{c}_k}^{\bm{\hat{q}_p}}, \cdots c_L] \\
    \hat{E} &= \{ \bm{\hat{q}}_p, \bm{l}_p \} \cup \{(\bm{q}_i, \bm{l}_i)\}_{i=1, i \neq p}^N
\end{align}
with $\bm{\hat{q}}_p = [\hat{c}_j, \cdots, \hat{c}_k]$ the new referring expression where the color attribute is changed. Specifically, we replace in $\bm{q}_p$ the color with a new randomly sampled one. Sampling is done on a vocabulary $C$ of 12 color attributes based on~\cite{DBLP:journals/tgrs/ZhanXY23}: black, gray, white, red, orange, yellow, green, cyan, blue, purple, pink, and brown. The variation function $\phi$ is applied 6 times ($|C| / 2$) per referring expression with color attribute. We chose 6 as a trade-off between the number of annotations generated and the variability introduced through multiple sampling.

The current definition of $\phi$ keeps fixed all objects' locations and $N - 1$ referring expressions. This is done to preserve the spatial arrangement of the objects, i.e. the layout and the image context. Objects’ locations are particularly relevant as they express complex semantic meaning. For example, the size of bounding boxes may express perspective: a person in the foreground should be bigger with respect to one in the background. Moreover, they could also identify relations: in the scene represented by ``a person reading a book'', the bounding box of the book should be small but also close to the bounding box of the person. In order to keep this rich semantic, in this work we prefer to focus on variation of text, which is more intuitive to generate and evaluate.

\subsection{The Image Generation Engine}
\label{sec:ige}

The Image Generation Engine (IGE) is responsible for generating synthetic images. This component receives an annotation $\bm{\hat{a}}$ obtained from the Annotation Generation Engine. It returns an image $I \in \mathcal{I}$ from the domain of images encoding semantic information expressed in $\bm{\hat{a}}$. More in detail, we define the IGE as a function $\psi: \mathcal{A} \rightarrow \mathcal{I}$: $\psi(\bm{\hat{a}}) = I$. 

We implement the Image Generation Engine component with Grounded-Language-to-Image Generation (GLIGEN)~\cite{DBLP:conf/cvpr/LiLWMYGLL23}. GLIGEN is a generative model based on Stable Diffusion~\cite{DBLP:conf/cvpr/RombachBLEO22}, which is capable of generating detailed and high-quality images. Although the pipeline does not bind the IGE component with a specific generative model, we chose GLIGEN for different reasons. Unlike mainstream generative models, GLIGEN allows fine-grained control over the output image. This is a fundamental aspect because we are interested in providing samples for REC task. Specifically, we are interested in generating images that are coherent to the annotations, i.e. locations of the referred objects. For this reason, a critical feature of the chosen generator is the ability to guide the synthesizing process through ``objects description'', beyond the image caption. That is, an image is generated by describing its content through a caption as in Stable Diffusion, but a set of pairs (referring expression, object location), namely entities, is also provided. These entities instruct GLIGEN with the objects' location and information on their appearance features. The more accurate the positioning of objects and fidelity to descriptions, the better the supervision signal for the Referring Expression Comprehension task.

Although the main focus of GLIGEN is the conditioning on entities, i.e., description and location of objects, it can also work with other modalities: images, keypoints, hed map, canny map, semantic map, normal map. Every modality can be used to control the generation of the output image. In this work, we focus on the standard modality, which is compatible with the format of annotation $\bm{\hat{a}}$ produced by the Annotation Generation Engine. 

\section{Harlequin Dataset}
\label{sec:dataset}

\begin{figure}[tb]
    \centering
    \includegraphics[width=1\linewidth]{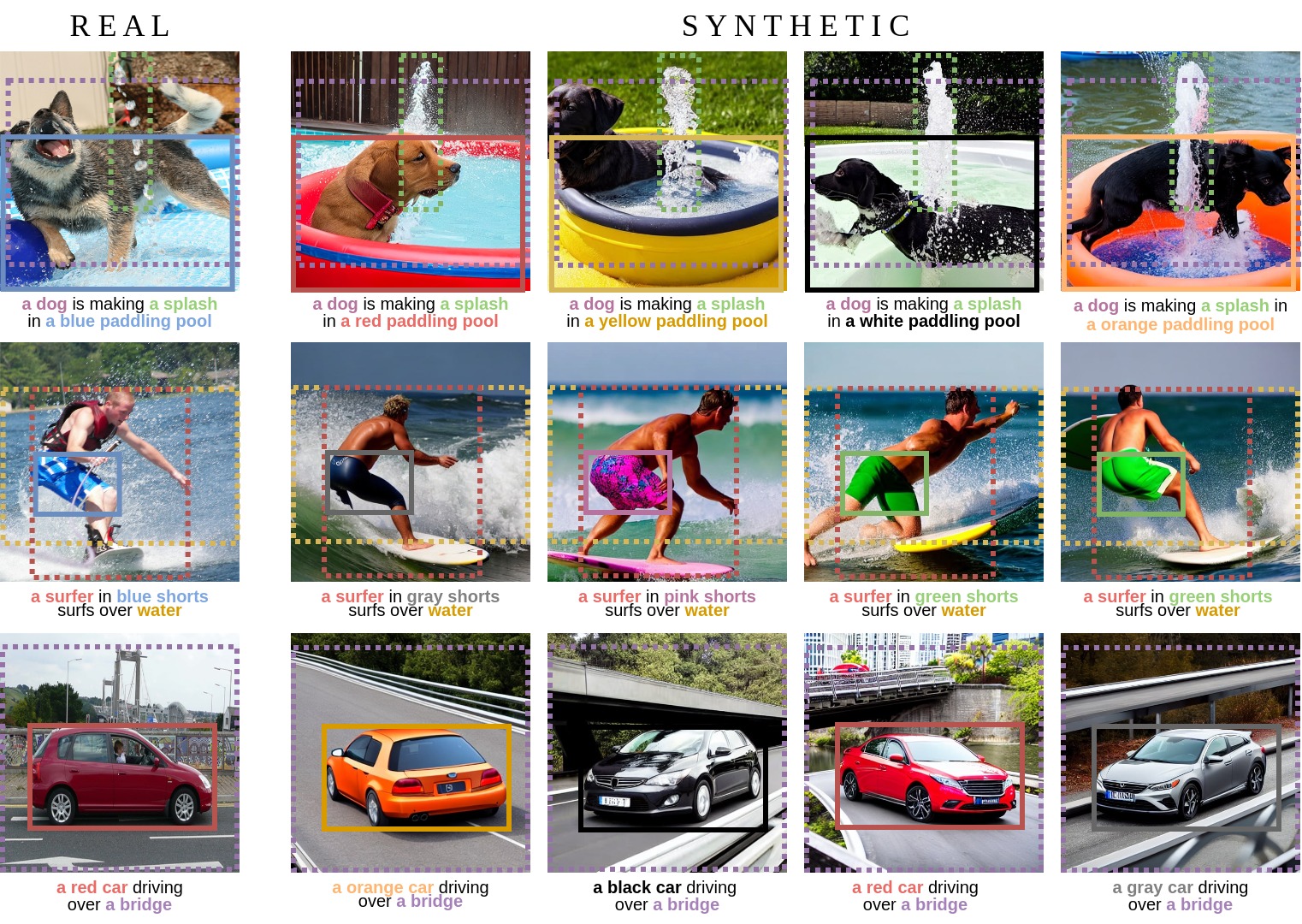}
    \caption{Examples produced by our pipeline. On the left, we show reference images along with their annotations from Flickr30k Entities. On the right, we report some generated variations. Colors are altered and guide, along with objects' locations, the image synthesis.}
    \label{fig:teaser}
\end{figure}

We introduce the first totally synthetic generated dataset for the Referring Expression Comprehension task, termed Harlequin,\footnote{Harlequin, or Arlecchino in Italian, is a character from the Italian commedia dell'arte known for his colorful patched costume.} collected via our pipeline. We report some examples in Fig.~\ref{fig:teaser}.
The dataset is originated from Flickr30k Entities: we select samples characterized by referring expressions containing the \textit{color} attribute to variate them.
Since the Image Generation Engine is eager in terms of resources, we first generate all the new annotations with Annotation Generation Engine using the selected samples as seeds. Secondly, we run the image synthesis adopting a frozen instance of GLIGEN in ``generation'' mode with ``text + box'' modality and batch size 1.\footnote{\href{https://github.com/gligen/GLIGEN}{\texttt{https://github.com/gligen/GLIGEN}}} 

\begin{figure*}[t]
  \centering
  \begin{subfigure}{0.49\linewidth}
    \centering
    \includegraphics[width=\linewidth]{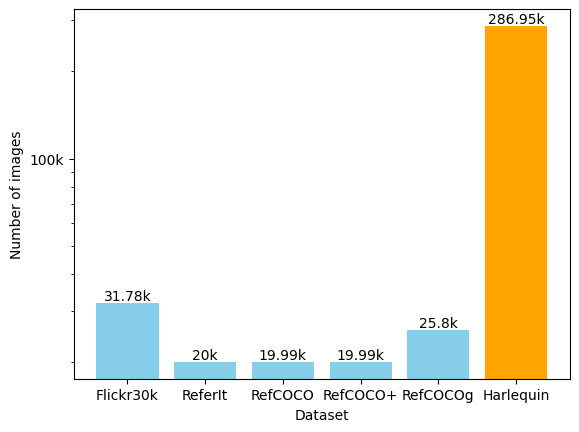}
  \end{subfigure}
  \hfill
  \begin{subfigure}{0.49\linewidth}
    \centering
    \includegraphics[width=\linewidth]{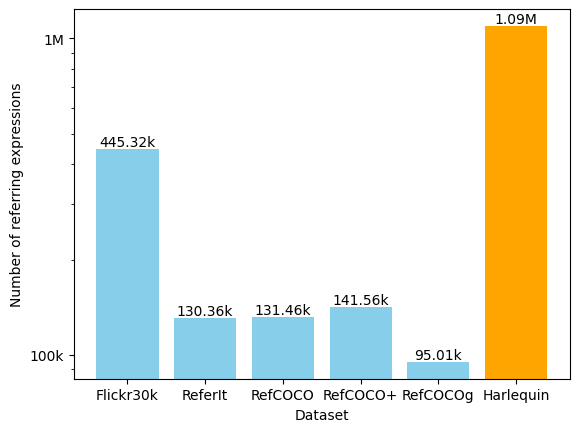}
  \end{subfigure}
  \caption{Dataset statistics. We report the number of images and referring expressions per dataset on the left and right, respectively. Harlequin is highlighted in orange.}
  \label{fig:dataset-stats}
\end{figure*}

Harlequin comprises a total of 286,948 synthetic images and 1,093,181 annotations targeting color attributes and following the coco format. It has 2.60$\pm$1.14 words per referring expression on average, in line with Flickr30k Entities statistics. The median value is 2, while the longest referring expression is 14 words. Harlequin follows Flickr30k Entities' data splits. It provides 988,342 annotations over 259,930 images for the training set, 52,554 annotations over 13,584 images for the validation set, and 52,284 annotations over 13,434 images for the test set. 
Fig.~\ref{fig:dataset-stats} visualizes the amount of data in Harlequin compared to existing, manually annotated, and collected datasets. Harlequin doubles the amount of referring expression in Flickr30k, the largest dataset available in the literature, and provides a noticeably larger amount of images.

Harlequin presents some interesting properties. For instance, the generated images display the same objects under various orientations and on different backgrounds, increasing the variability and complexity of Harlequin with respect to Flickr30k Entities, while retaining its supervision signal (Fig.~\ref{fig:teaser}, third row). Moreover, we observe that our generation strategy fixes some errors in the human-annotated labels. As a matter of fact, we noticed that Flickr30k Entities contains some samples annotated with the wrong locations of the bounding boxes. The pipeline addresses this issue, generating new images coherent with the given annotations where the referred object is correctly inside the provided bounding box. 
Finally, we bring up that the used variation function $\phi$ inevitably leads to the generation of unrealistic-colored objects (e.g. ``the blue dog''). Independently of that, the results show that Referring Expression Comprehension models learn a robust representation from Harlequin. This is coherent with the fact that humans are usually capable of identifying an object regardless of its color and use this information to disambiguate similar objects. 

\section{Experiments}

We present experimental results obtained in Referring Expression Comprehension by pre-training two models on Harlequin, our synthetic dataset, and fine-tuning them on realistic datasets. We show that the pre-training improves performance. We discuss the role that variations in original annotation play in the improvement of results, and finally, we analytically evaluate the contribution of the color variations through an ablation study.

\subsection{Implementation details}
\label{sec:impl}

We pre-train TransVG~\cite{DBLP:conf/iccv/DengYCZL21}, VLTVG~\cite{DBLP:conf/cvpr/YangXYLLH22}, and LGR-NET~\cite{DBLP:journal/tcsv/LuLFMW24} on our synthetic dataset and then fine-tune them on RefCOCO family datasets initializing the weights of the model with those obtained after the pre-training. We followed the implementation details of TransVG, VLTVG, and LGR-NET. For all the experiments on TransVG and VLTVG, \textit{pre-training} on Harlequin, \textit{fine-tuning} on RefCOCO family, and from-scratch \textit{baselines}, we initialized the weights of the visual transformers with those of DETR~\cite{DBLP:conf/eccv/CarionMSUKZ20} based on the ResNet-50~\cite{DBLP:conf/cvpr/HeZRS16} available on projects' page. Instead, the linguistic branch is initialized with the weights of the BERT model~\cite{DBLP:conf/naacl/DevlinCLT19}. We set the batch size to 32 and use AdamW as the optimizer. For the \textit{pre-training} on Harlequin, we train both TransVG and VLTVG for 60 epochs, the value suggested by authors for Flickr30k, dropping the learning rate by a factor of 10 after 40 epochs. Instead, the \textit{fine-tuning} experiments and supervised \textit{baselines} are trained for 90 epochs with a learning rate dropped by a factor of 10 after 60 epochs. When using the TransVG model, we set the weight decay to $10^{-4}$, the initial learning rate of the vision-language module and prediction head to $10^{-4}$, and of the visual branch and linguistic branch to $10^{-5}$. When using the VLTVG model, the initial learning rate of the feature extraction branches is $10^{-5}$ and $10^{-4}$ for all the other components. Moreover, we freeze the weights of the visual and textual branches in the first 10 epochs. As concerns LGR-NET model, we pre-trained the model on Harlequin and the fine-tuning and supervised baselines experiments are trained for 15 epochs. We used Swin Transformer Small~\cite{DBLP:conf/iccv/LiuL00W0LG21} as the backbone, BERT as the textual extractor, and followed the implementation details provided by the authors. During evaluation, we set batch size to 32. We carried out all the experiments on a single NVIDIA RTX A5000.
We used the code provided online.\footnote{\href{https://github.com/djiajunustc/TransVG}{\texttt{https://github.com/djiajunustc/TransVG}}}\textsuperscript{,}\footnote{\href{https://github.com/yangli18/VLTVG}{\texttt{https://github.com/yangli18/VLTVG}}}\textsuperscript{,}\footnote{\href{https://github.com/lmc8133/LGR-NET}{\texttt{https://github.com/lmc8133/LGR-NET}}}

\subsection{Evaluation Models and Metrics}

The models we chose for the evaluation of Harlequin on the Referring Expression Comprehension task are TransVG~\cite{DBLP:conf/iccv/DengYCZL21}, VLTVG~\cite{DBLP:conf/cvpr/YangXYLLH22}, and LGR-NET~\cite{DBLP:journal/tcsv/LuLFMW24}, as mentioned above. TransVG proposes an alternative prediction paradigm to directly regress the target coordinates. It makes use of transformer for both intra- and inter-modality correspondence. A regression token is added to the multi-modal transformer and is optimized through a regression head that directly outputs the object's location. VLTVG employs a visual-linguistic verification mechanism alongside a language-guided context encoder to extract discriminative features of the referred object. The visual-linguistic verification module enhances visual features, emphasizing regions related to the referring expression, whereas the language-guided context encoder collects meaningful visual contexts. Ultimately, a multi-stage cross-modal decoder is utilized to iteratively analyze the encoded visual and textual features, refining the object representation for precise target localization. LGR-NET emphasizes the guidance of textual features for cross-modal reasoning extending the standard textual features generating three embeddings: coordinate, word, and sentence. The textual features are, then, employed for alternated cross-modal reasoning exploiting a loss enhances the cross-modal alignment while localizing the referred object.

The evaluation metric is the standard accuracy. Given a referring expression, it considers a prediction to be correct if and only if the intersection over union between the predicted and the ground truth bounding box is at least $0.5$.

\subsection{Results}
\label{sec:results}

\begin{table}[tbp]
    \centering
    \begin{tabular}{l|ccc|ccc|cc}
        \toprule
        \multirow{2}{*}{\textbf{Method}} & \multicolumn{3}{c|}{\textbf{RefCOCO}} & \multicolumn{3}{c|}{\textbf{RefCOCO+}} &  \multicolumn{2}{c}{\textbf{RefCOCOg}} \\
         & val & testA & testB & val & testA & testB & val & test \\
        \midrule
        \multicolumn{9}{l}{\textit{\textbf{TransVG~\cite{DBLP:conf/iccv/DengYCZL21}}:}} \tstrut\bstrut\\
        Real & 63.33 & 69.05 & 55.62 & 64.69 & 69.02 & \textbf{55.76} & 64.04 & 63.22 \tstrut\\
        Synth$\rightarrow$Real & \textbf{65.77} & \textbf{70.66} & \textbf{56.80} & \textbf{66.66} & \textbf{72.01} & 55.66 & \textbf{65.13} & \textbf{64.33} \\ 
        \small{(Improv.)} & \improv{+2.44} & \improv{+1.61} & \improv{+1.18} & \improv{+1.97} & \improv{+2.99} & \small\textcolor{gray}{-0.10} & \improv{+1.09} & \improv{+1.11} \\
        \midrule
        \multicolumn{9}{l}{\textit{\textbf{VLTVG~\cite{DBLP:conf/cvpr/YangXYLLH22}}:}} \tstrut\bstrut\\
        Real & \textbf{69.66} & 74.33 & \textbf{61.35} & 70.83 & 76.02 & \textbf{61.71} & \textbf{70.57} & \textbf{70.03} \tstrut\\
        Synth$\rightarrow$Real & 69.60 & \textbf{75.76} & 61.14 & \textbf{71.46} & \textbf{77.16} & 61.30 & 70.04 & 69.57 \\
        \small{(Improv.)} & \small\textcolor{gray}{-0.06} & \improv{+1.43} & \small\textcolor{gray}{-0.21} & \improv{+0.63} & \improv{+1.12} & \small\textcolor{gray}{-0.41} & \small\textcolor{gray}{-0.53} & \small\textcolor{gray}{-0.46} \\
        \midrule
        \multicolumn{9}{l}{\textit{\textbf{LGR-NET~\cite{DBLP:journal/tcsv/LuLFMW24}}:}} \tstrut\bstrut\\
        Real & 82.71 & 85.77 & 79.31 & 71.11 & 75.45 & 63.35 & 70.75 & 71.11 \tstrut\\
        Synth$\rightarrow$Real & \textbf{84.38} & \textbf{87.13} & \textbf{80.67} & \textbf{71.40} & \textbf{75.60} & \textbf{64.70} & \textbf{74.61} & \textbf{75.22} \\
        \small{(Improv.)} & \improv{+1.67} & \improv{+1.36} & \improv{+1.36} & \improv{+0.29} & \improv{+0.15} & \improv{+1.35} & \improv{+3.86} & \improv{+4.11} \\
        \bottomrule
    \end{tabular}
    \caption{Results. We show the performance of three methods, TransVG, VLTVG and LGR-NET, on the Referring Expression Comprehension task with pre-training on Harlequin (\textit{Synth$\rightarrow$Real}) and without (\textit{Real}). The pre-training shows superior or comparable performance on three benchmarks: RefCOCO, RefCOCO+ and RefCOCOg. We report the standard accuracy percentage.}
    \label{tab:results}
\end{table}

Tab.~\ref{tab:results} shows the performance of TransVG~\cite{DBLP:conf/iccv/DengYCZL21}, VLTVG~\cite{DBLP:conf/cvpr/YangXYLLH22}, and LGR-NET~\cite{DBLP:journal/tcsv/LuLFMW24} in Referring Expression Comprehension task. Specifically, we report the results obtained in two settings. In the first, we train the model from scratch on realistic datasets: RefCOCO, RefCOCO+, and RefCOCOg. In the second, we pre-train the model on Harlequin, our synthetic dataset, and then fine-tune it on realistic datasets. In both cases, we report the evaluation on the three RefCOCO datasets.

TransVG shows homogeneous improvement among all datasets. It improves by $2.44\%$, $1.61\%$, and $1.18\%$ in RefCOCO splits and shows superior performance also in RefCOCO+ and RefCOCOg. For VLTVG, despite the model starts from a higher performance with respect to TransVG, it shows a remarkable $1.43\%$ and $1.12\%$ improvement on RefCOCO and RefCOCO+'s testA. As concerns LGR-NET, we improve the supervised baselines on all the datasets reaching up to $+3.86\%$ and $+4.11\%$ on the RefCOCOg splits. We recall that the reported improvement emerges in a cross-dataset setting. As a matter of fact there is no overlap between the pre-training data, synthetically generated from Flickr30k Entities, and the fine-tuning datasets.

The results demonstrate that pre-training on synthetically generated data is feasible in the Referring Expression Comprehension task. Annotations required by this task challenge generative models, where their artistic traits need to deal with fine-grained constraints on objects' locations and descriptions. Nevertheless, our pipeline proves that the generation and collection of heavily annotated data with zero human effort is possible. This is an important milestone and opens a wide range of future directions where data can be crafted to overcome the increasing need for annotations. 
We argue that the artificial nature of data is overcome when the control over semantic properties is appropriately exploited. Merely generating a dataset may not imply good performance, especially if the model is tested on realistic benchmarks. The generated dataset must encode some knowledge that the model can learn in order to compete with real-world datasets.

\subsection{Impact of the Color Attribute}

\begin{table}[t]
    \centering
    \begin{tabular}{llc|cc|cc}
        \toprule
         \multicolumn{3}{c|}{\textbf{Evaluation}} & \multicolumn{2}{c|}{\textbf{TransVG~\cite{DBLP:conf/iccv/DengYCZL21}}} & \multicolumn{2}{c}{\textbf{VLTVG~\cite{DBLP:conf/cvpr/YangXYLLH22}}} \\
        Dataset &  & \% Anns & Real & Synth$\rightarrow$Real & Real & Synth$\rightarrow$Real \\
        \midrule
        \multirow{3}{*}{RefCOCO} & val (color) & 23.2 & 64.35 & \textbf{69.50} \improv{(+5.15)} & \textbf{77.15} & 77.11 \small\textcolor{gray}{(-0.04)} \\
         & testA (color) & 37.5 & 68.73 & \textbf{72.91} \improv{(+4.18)} & 79.46 & \textbf{81.48} \improv{(+2.02)} \\
         & testB (color) & 17.8 & 56.46 & \textbf{60.11} \improv{(+3.65)} & 68.95 & \textbf{70.28} \improv{(+1.33)} \bstrut\\
        \hline \tstrut
        \multirow{3}{*}{RefCOCO+} & val (color) & 34.6 & 68.91 & \textbf{70.07} \improv{(+1.16)} & 75.79 & \textbf{77.13} \improv{(+1.34)} \\
         & testA (color) & 37.5 & 71.63 & \textbf{74.20} \improv{(+2.59)} & 79.13 & \textbf{80.95} \improv{(+1.82)} \\
         & testB (color) & 26.8 & 55.73 & \textbf{57.33} \improv{(+1.60)} & \textbf{65.65} & 63.82 \small\textcolor{gray}{(-1.83)} \bstrut\\
        \hline \tstrut
        \multirow{2}{*}{RefCOCOg} & val (color) & 41.4 & 62.37 & \textbf{65.04} \improv{(+2.67)} & \textbf{73.23} & 73.09 \small\textcolor{gray}{(-0.14)} \\
         & test (color) & 41.7 & 62.12 & \textbf{64.94} \improv{(+2.82)} & 73.05 & \textbf{73.70} \improv{(+0.65)} \\
        \bottomrule
    \end{tabular}
    \caption{Ablation study. We evaluate the performance of TransVG and VLTVG on a subset of test sets where referring expressions contain at least one color attribute. Column \textit{\% Anns} reports the percentage of annotations with at least a color attribute with respect to original test sets. Columns \textit{Real} and \textit{Synth$\rightarrow$Real} show the performance without or with pre-training on Harlequin. We report standard accuracy in percentage.}
    \label{tab:ablation}
\end{table}

In this section, we evaluate the impact of our pre-training on realistic samples with the \textit{color} attribute. We follow the same training scheme. However, here the test sets are limited to samples containing a referring expression with a color. As shown in Tab.~\ref{tab:ablation}, TransVG~\cite{DBLP:conf/iccv/DengYCZL21} demonstrates a boost in performance among RefCOCO family datasets, with remarkable $+5.15\%$, $+4.18\%$ and $+3.65\%$ on RefCOCO. The pre-training shows superior or comparable performance also for VLTVG~\cite{DBLP:conf/cvpr/YangXYLLH22}, with the exception of testB for RefCOCO+. We recall that no changes to the model's architecture have been made to encode extra knowledge about colors. The improvement is solely guided by learning patterns from data.

However, these results were expected. As a matter of fact, Harlequin is mainly composed of referring expressions that contain a color attribute. Consequently, models pre-trained on our dataset primarily acquire generalization capabilities in identifying and distinguishing objects with different colors.

\section{Conclusion}
\label{sec:conclusion}

In this work, we design a new pipeline that aims to generate synthetic data for the Referring Expression Comprehension task. It involves two components: the Annotation Generation Engine for creating new expressive annotations and the Image Generation Engine to generate synthetic images conditioned by the annotations. Our strategy can generate datasets with arbitrary dimensions and complexity without human effort and reduce some errors in labeling. We adopt the method to generate Harlequin, the first dataset collected for the Referring Expression Comprehension task. Harlequin is built on top of Flickr30k Entities' annotations and is generated varying color attributes in the original referring expressions. We validate our approach by pre-training state-of-the-art models on Harlequin and demonstrate that the acquired generalization capabilities improve the performance after the fine-tuning on real data.

In future work, we plan to investigate the potential and flexibility of our pipeline to progressively get rid of each input until the entire sample is generated from scratch. In particular, we believe that the proposed variation function could be extended to work with other attributes besides the color or could be learned. Some of them, such as \textit{size} and \textit{location}, also require the manipulation of bounding boxes' coordinates besides queries. There has been effort to face this new challenge. For example, LayoutGPT~\cite{DBLP:journals/corr/abs-2305-15393} generates a reasonable arrangement of objects given a textual description and returns their coordinates. This tool, combined with our pipeline, could alleviate the problem of having fixed layout of objects. Finally, we believe that the generation of the referring expressions could be automatized through prompting strategies, which have been proven effective for task adaptation in Large Language Models~\cite{DBLP:conf/iclr/AroraNCOGBCR23}.

\bigskip\paragraph{\textbf{Acknowledgments.}}
We acknowledge the CINECA award under the ISCRA initiative, for the availability of high performance computing resources and support.
This work is also supported by the PNRR project FAIR - Future AI Research (PE00000013), under the NRRP MUR program funded by NextGenerationEU.

%
%

{
    \small
    \bibliographystyle{splncs04}
    \bibliography{main}
}

\end{document}